\pdfoutput=1

\documentclass[11pt]{article}

\usepackage{ACL2025}

\usepackage{times}
\usepackage{latexsym}

\usepackage{multirow} 
\usepackage{adjustbox} 
\usepackage{caption} 
\usepackage{longtable}
\usepackage{booktabs} 
\usepackage{colortbl} 
\usepackage{subcaption}
\usepackage{natbib}
\usepackage{algorithm}
\usepackage{algpseudocode}
\usepackage{makecell} 
\usepackage{kotex}
\usepackage{tcolorbox}
\usepackage{amsmath}
\usepackage{enumitem}

\usepackage[T1]{fontenc}

\usepackage[utf8]{inputenc}

\usepackage{microtype}

\usepackage{inconsolata}

\title{FEAT: A Preference Feedback Dataset through a Cost-Effective Auto-Generation and Labeling Framework for English AI Tutoring}

\author{
Hyein Seo$^1$, Taewook Hwang$^1$, Yohan Lee$^2$, Sangkeun Jung$^1$$^3$\thanks{\ \ Corresponding author} \\
$^1$Chungnam National University \\
$^2$Electronics and Telecommunications Research Institute \\
$^3$EurekaAI\\
\texttt{\{hyenee97,taewook5295\}@gmail.com},
\texttt{carep@etri.re.kr},
\texttt{hugmanskj@gmail.com}
}

\begin{document}
\maketitle
\begin{abstract}
In English education tutoring, teacher feedback is essential for guiding students. Recently, AI-based tutoring systems have emerged to assist teachers; however, these systems require high-quality and large-scale teacher feedback data, which is both time-consuming and costly to generate manually. In this study, we propose FEAT, a cost-effective framework for generating teacher feedback, and have constructed three complementary datasets\footnote{Our dataset is publicly available at \url{https://github.com/hyenee/FEAT}}: (1) DIRECT-Manual (DM), where both humans and large language models (LLMs) collaboratively generate high-quality teacher feedback, albeit at a higher cost; (2) DIRECT-Generated (DG), an LLM-only generated, cost-effective dataset with lower quality;, and (3) DIRECT-Augmented (DA), primarily based on DG with a small portion of DM added to enhance quality while maintaining cost-efficiency. Experimental results showed that incorporating a small portion of DM (5–10\%) into DG leads to superior performance compared to using 100\% DM alone.
\end{abstract}

\section{Introduction}\label{sec:introduction}
In English education tutoring, providing appropriate teacher feedback plays a crucial role in guiding students and improving their educational outcomes \cite{its,positive_feedback_its}. Given its importance, various studies have explored automated teacher feedback generation \cite{feedback_generation1,feedback_generation2,feedback_generation3}.

\begin{figure}
   \centering
   \includegraphics[width=0.95\linewidth]{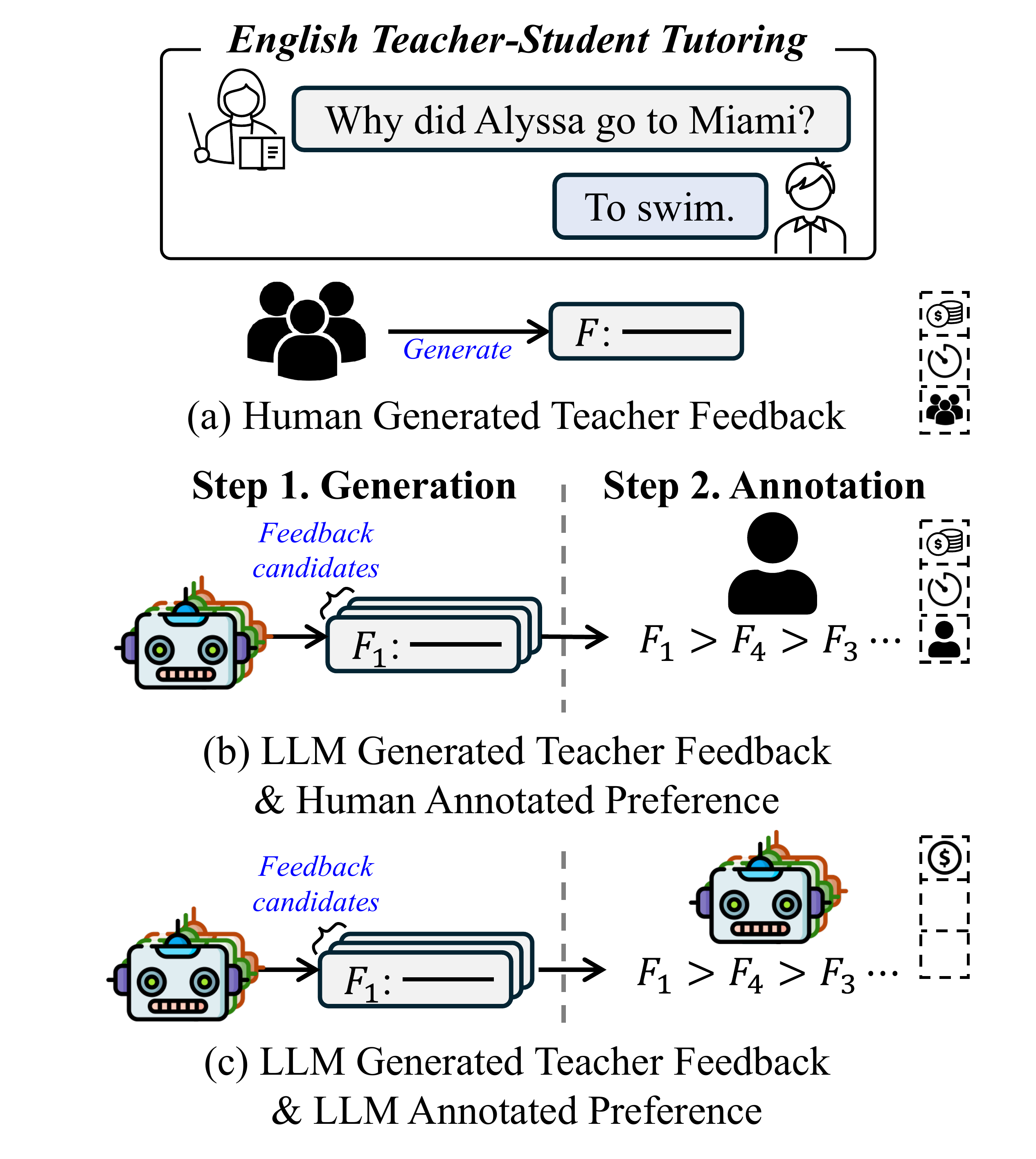}
   \vspace{-5mm}
   \caption{Teacher feedback generation and annotation process in an English tutoring system.}
   \label{fig:overview}
\end{figure}

Figure \ref{fig:overview} illustrates methods for generating and annotating teacher feedback for tutoring systems. As shown in (a), human-generated feedback with ranking provides high quality, but its time-consuming and costly nature makes it difficult to scale up \cite{data_annotation}.

To address this challenge, we propose \textbf{FEAT}, a cost-effective framework using large language models (LLMs) to automatically generate a large-scale teacher feedback preference dataset for tutoring AI. This enables reward- or rank-based learning, making it suitable for building human-friendly tutoring models \cite{rlhf}. FEAT generates teacher feedback based on student responses, using the dialogue history between teacher and student and context as input. Moreover, we apply feedback criteria defined by \citet{hyein-llm-evaluators} to ensure educationally appropriate feedback.

\begin{figure*}
    \centering
    \includegraphics[width=1.0\linewidth]{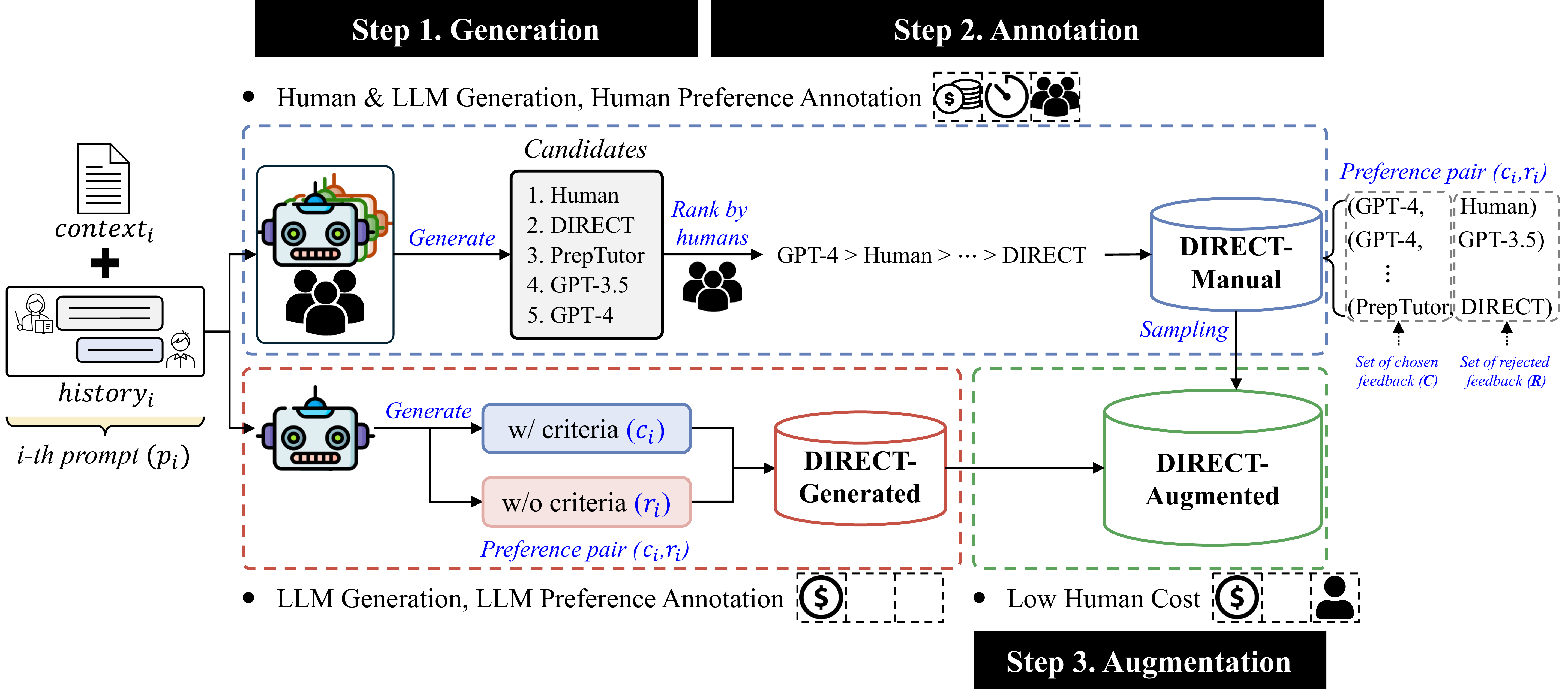}
    \caption{The architecture of the FEAT framework, illustrating the construction process of the DIRECT-Manual, DIRECT-Generated, and DIRECT-Augmented datasets. $p_{i}$, $c_{i}$, and $r_{i}$ denote the $i$-th prompt, chosen, and rejected responses, respectively.}
    \label{fig:architecture}
\end{figure*}

Using FEAT, we constructed three datasets: (1) \textit{DIRECT-Manual} (DM), which contains human and LLM-generated feedback with human-annotated rankings  (high quality and high cost), (2) \textit{DIRECT-Generated} (DG), an entirely LLM-generated and annotated preference dataset (medium quality and low cost), and (3) \textit{DIRECT-Augmented} (DA), a hybrid dataset built on DG with a minor addition of DM (high quality and low cost).

Our experiments showed that incorporating a small portion of DM (5–10\%) into DG leads to superior performance compared to using DM alone.

Our main contributions are as follows: 

\begin{itemize}[leftmargin=15pt, itemsep=0pt, parsep=0pt]
    \item We proposed FEAT, a cost-effective framework for automated teacher feedback generation and annotation in English tutoring.
    \item We constructed three preference datasets: DM, DG, and DA, enabling reward- or rank-based learning.
    \item We confirmed that incorporating a small amount of DM into DG (DA) yielded better performance than DM alone.
\end{itemize}

\section{FEAT: \textit{F}eedback Dataset Generation Framework for \textit{E}nglish \textit{A}I \textit{T}utoring}\label{sec:feedback_dataset}
Figure \ref{fig:architecture} illustrates the construction process of our FEAT framework.
FEAT applies five criteria from \citet{hyein-llm-evaluators}-\textit{Correct, Revealing, Guidance, Diagnostic,} and \textit{Encouragement}—ensuring educationally effective feedback.

\subsection{DIRECT-Manual: Rank-based Preference Dataset}\label{sec:direct_m}

DM is an extended version of the DIRECT \cite{direct} dataset, simulating intelligent tutoring between teachers and students. While it ensures high quality, it relies heavily on human effort, making it time-consuming and costly.

\begin{figure}
    \centering
    \includegraphics[width=\linewidth]{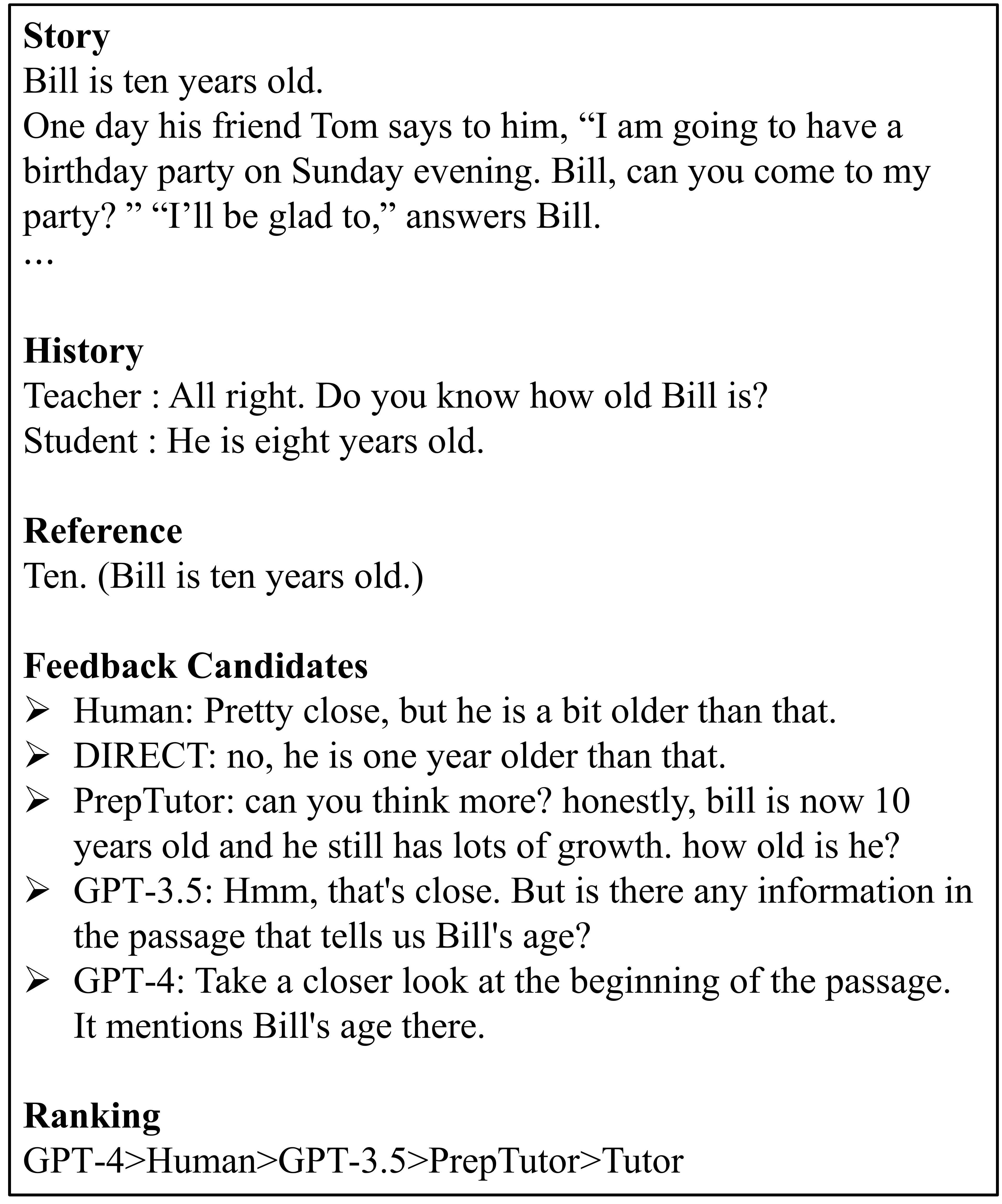}
    \caption{Sample from the DIRECT-Manual.}
    \label{fig:sample_direct}
\end{figure}

\noindent\textbf{Step 1: Feedback Generation.}
We collected teacher feedback data for scenarios with incorrect student answers in teacher-student dialogues from diverse sources (Human, DIRECT, PrepTutor, GPT-3.5, and GPT-4; see Appendix \ref{appendix:details-of-direct_m}), with DM previously used as private data in \citet{direct-dpo}. An example from the DM is shown in Figure \ref{fig:sample_direct}. 

\noindent\textbf{Step 2: Feedback Ranking via Human Annotation.}
Human annotators ranked the five feedback candidates using two criteria: \textit{Correct} (specific and accurate information) and \textit{Revealing} (avoiding direct answers). Feedback meeting both criteria received the highest rank, with Correct prioritized when only one criterion was met. 

\noindent\textbf{Step 3: Preference Data Construction.}
From ranked five feedback candidates, we created pairwise combinations; in each pair, the higher-ranked feedback is labeled \textit{\textbf{chosen}} and the lower-ranked as \textit{\textbf{rejected}}.

\subsection{DIRECT-Generated: Criteria-based Preference Dataset}\label{sec:direct_a}

DG uses LLM to automatically generate and annotate teacher feedback based on specific criteria, producing reasonably good data at a lower cost.

\begin{figure}
    \centering
    \includegraphics[width=\linewidth]{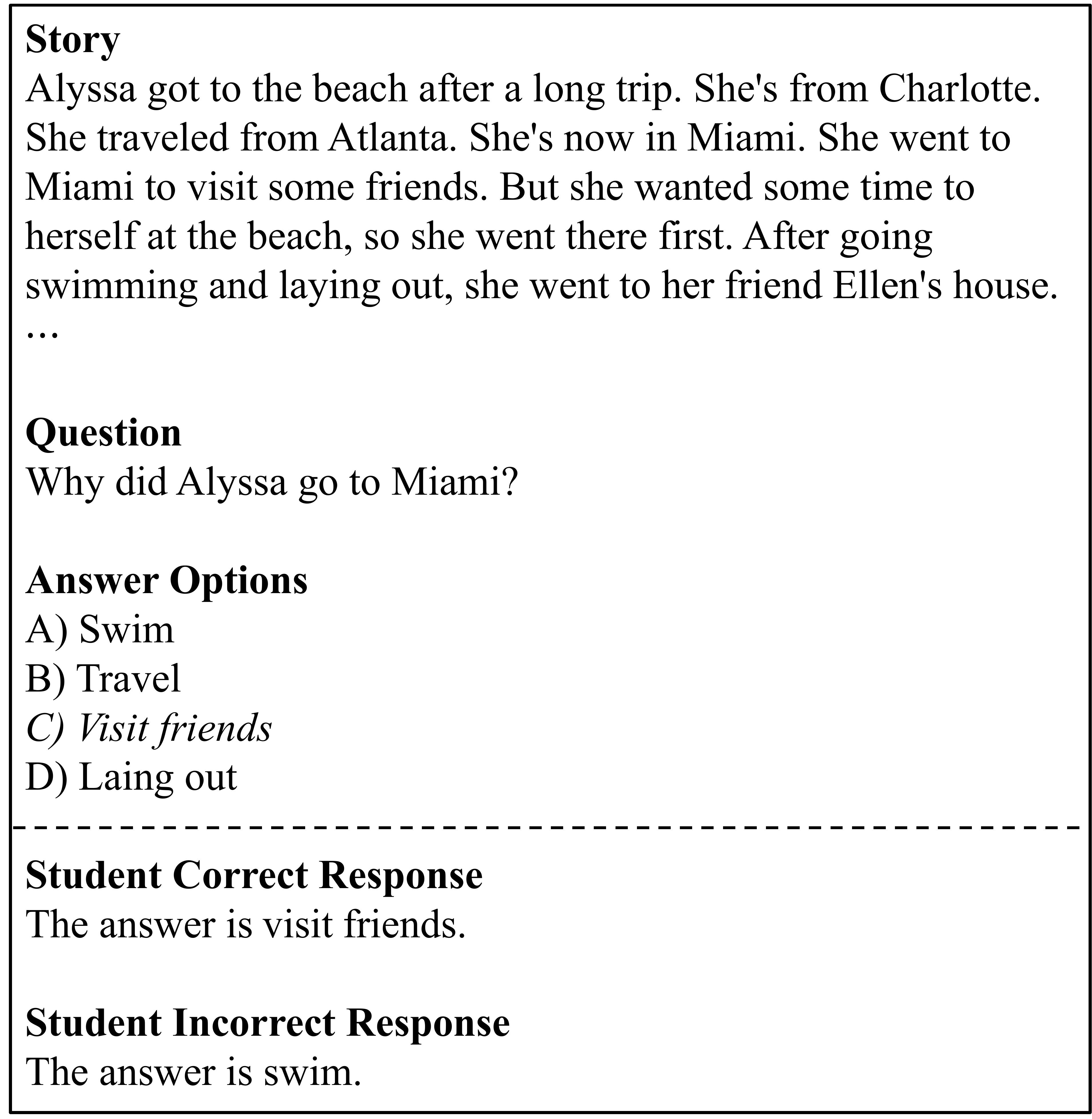}
    \caption{Sample from the MCTest.}
    \label{fig:sample_mctest}
\end{figure}

\noindent\textbf{Step 1: Feedback Generation.}
Using dialogue history and context, LLM generates teacher feedback based on five criteria. We created tutoring scenarios by converting reading comprehension tasks from MCTest \cite{mctest} to generate large-scale feedback data. A sample from the MCTest is illustrated in Figure \ref{fig:sample_mctest}.

\noindent\textbf{Step 2: Preference Data Construction.}
We generated two types of feedback: \textit{w/ criteria} (applying five criteria) and \textit{w/o criteria} (without criteria). We labeled \textit{w/ criteria} feedback as \textit{\textbf{chosen}} and \textit{w/o criteria} as \textit{\textbf{rejected}}, assuming criteria-based feedback is of higher quality. Data statistics are shown in Table \ref{tab:dataset_summary}, with full details in Appendix \ref{appendix:details-of-direct_a}.

\begin{table}
\centering
\renewcommand{\arraystretch}{0.8}
\begin{adjustbox}{max width=\linewidth}
\begin{tabular}{ccc}
\toprule 
 & Train & Test\tabularnewline
\midrule 
DIRECT-Manual & 5,025 & 475\tabularnewline
DIRECT-Generated & 3,996 & 444\tabularnewline
\bottomrule
\end{tabular}
\end{adjustbox}
\caption{\label{tab:dataset_summary}Dataset statistics.}
\end{table}

\begin{figure*}
   \centering
   \includegraphics[width=\linewidth]{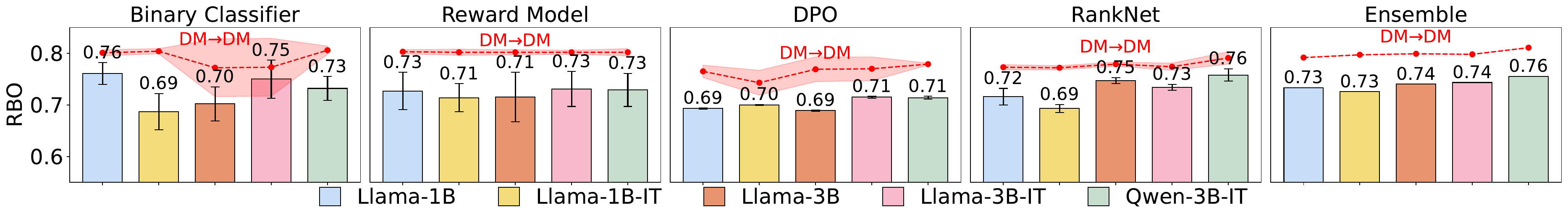}
\caption{Ranking model performance across different approaches (with 5-seed standard deviation). Lines indicate DM$\rightarrow$DM performance, while bars show DG$\rightarrow$DM performance.}
   \label{fig:overall_performance}
\end{figure*}

\begin{figure*}
    \centering
    \subfloat[Llama-3B-IT\label{fig:performance_ratio_llama3B-IT_filtered}]{
        \includegraphics[width=\linewidth]{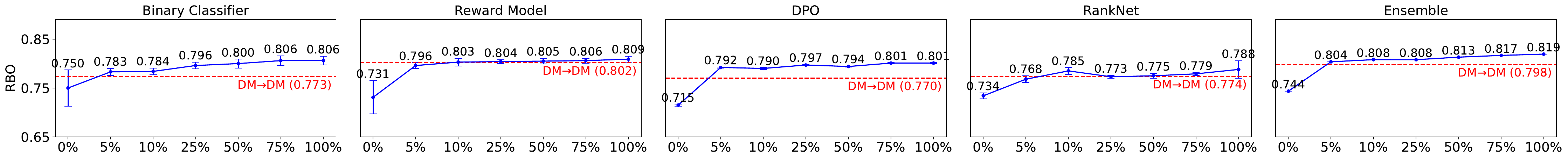}}

    \subfloat[Qwen-3B-IT\label{fig:performance_ratio_qwen3B-IT_filtered}]{
        \includegraphics[width=\linewidth]{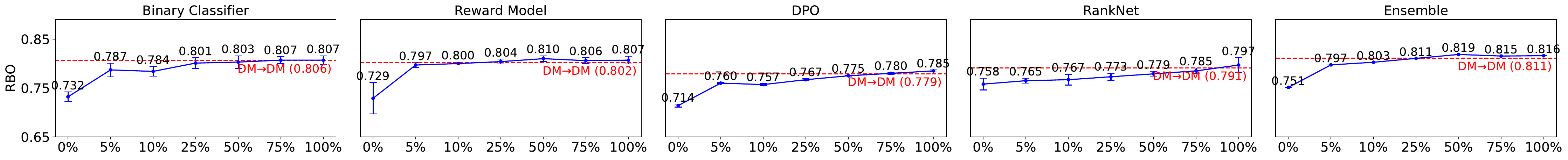}}

    \caption{\label{fig:performance_ratio_filtered} Llama-3B-IT and Qwen-3B-IT performance in the DA$\rightarrow$DM scenario. (See Appendix \ref{appendix:additional_experimental-results} for other models.)}
\end{figure*}

\section{Teacher Feedback Ranking}\label{sec:teacher_feedback_ranking}
To validate the preference annotations in DM, DG, and DA, we trained pairwise-based ranking models.

\subsection{Ranking Models}\label{sec:ranking_model}
We employed five approaches to train ranking models. Each model takes \textit{(prompt, chosen, rejected)} as input.

\noindent\textbf{Binary Classifier}
formulates preference learning as a binary classification task, labeling \textit{(chosen, rejected)} pairs as 1 and \textit{(rejected, chosen)} pairs as 0. The input sequence is depicted in Figure \ref{fig:input_sequence_binary_classifier}.

\noindent\textbf{Reward Model \cite{rlhf}}
computes scalar preference scores for feedback pairs, training to assign higher scores to chosen feedback.

\noindent\textbf{Direct Preference Optimization (DPO) \cite{dpo}}
optimizes language model probabilities to prefer chosen feedback, using log probability differences between chosen and rejected pairs.

\noindent\textbf{RankNet \cite{ranknet}}
learns score differences between feedback pairs using Binary Cross-Entropy loss. Reward Model, DPO, and RankNet share the same prompt, shown in Figure \ref{fig:prompt_preference_learning}.

\noindent\textbf{Ensemble}
aggregates predictions from the above four approaches through majority voting.

\subsection{Scenarios for Training}\label{sec:teacher_feedback_ranking:training_scenario}
We evaluated our ranking models on DM with three training configurations. The arrow ($\rightarrow$) indicates training (left) and evaluation (right).

\begin{tcolorbox}[colframe=black, colback=gray!10, rounded corners, left=-7pt]
\begin{itemize}
    \setlength{\itemsep}{0pt}  
    \setlength{\parsep}{0pt}   
    \item \textbf{DM$\rightarrow$DM}: Training with DM using manual annotation, serving as a performance upper bound for comparison with the other two scenarios (DG$\rightarrow$DM and DA$\rightarrow$DM).
    \item \textbf{DG$\rightarrow$DM}: Training with DG using automatic annotation.
    \item \textbf{DA$\rightarrow$DM}: Hybrid training using DG combined with a subset of DM for mixed annotation.
\end{itemize}
\end{tcolorbox}

During training, we enhanced data diversity by including feedback from different contexts beyond the standard \textit{(chosen, rejected)} pairs. This approach enabled the model to learn feedback comparisons across various contexts.

For evaluation, we tested the model on all possible pairs in DM. The model's pairwise predictions were aggregated to create overall rankings, with accuracy scored as 1 for \textit{chosen > rejected} and 0 for \textit{chosen < rejected}.

\section{Experiments}\label{sec:experiments}

We designed experiments and analyzed results to address the following research questions:
\begin{itemize}[leftmargin=15pt, itemsep=0pt]
    \item How does ranking model performance with DG compare to human-curated DM (Section \ref{sec:experimental_results:main_results})? 
    \item How does the ratio of DM in DA affect ranking model performance (Section \ref{sec:experimental_results:ratio})?
    \item How does the number of criteria in DG affect ranking models performance? (Section \ref{sec:experimental_results:compare_feedback_criteria})?
\end{itemize}

\subsection{Experimental Setup}\label{sec:experimental_setup}

\noindent\textbf{Models} 
We trained ranking models using five open-source models: Llama-1B \cite{llama3}, Llama-1B-IT, Llama-3B, Llama-3B-IT, and Qwen-3B-IT \cite{qwen}. Model details and hyperparameters are provided in Appendix \ref{appendix:implementation-details}.

\noindent\textbf{Evaluation Metrics} 
Rank-biased overlap (RBO) is a metric used to measure the similarity between two ranked lists. It ranges from 0 to 1, with values closer to 1 indicating higher similarity between the lists.

\subsection{Comparison of Ranking Model Performance}\label{sec:experimental_results:main_results}
As shown in Figure \ref{fig:overall_performance}, DM$\rightarrow$DM performed consistently (0.77-0.80) across all sizes and approaches. While DG$\rightarrow$DM showed lower but competitive results: the binary classifier reached 0.76 (Llama-1B), reward model 0.73 (Llama-3B-IT), DPO 0.71 (Llama-3B-IT, Qwen-3B-IT), RankNet 0.76 (Qwen-3B-IT), and Ensemble 0.76 (Qwen-3B-IT). Notably, the Ensemble maintained stable performance across different architectures, mitigating the variability seen in individual approaches.

These results indicated that teacher feedback generated by LLMs can produce rankings highly comparable to human annotator rankings, with a particularly strong trend observed in larger models. 

\noindent \textbf{Case Study}
As a result of analyzing the RBO scores between the ground-truth and predicted rankings, Figure \ref{fig:case_study_best} demonstrates that the two rankings are nearly identical, except for the swapped positions of DIRECT and PrepTutor, resulting in an RBO score of 0.8833. In contrast, Figure \ref{fig:case_study_worst} exhibits significantly lower agreement between the ground-truth and predicted rankings, with an RBO score of 0.4166. In DM, which is limited to five feedback candidates, the RBO score maintains a baseline similarity of at least 0.4 even when the rankings are completely different, due to the limited number of possible permutations. Additional experimental results are provided in Appendix \ref{appendix:additional_experimental-results}.

\begin{figure*}
    \centering
    \subfloat[Llama-1B\label{fig:criteria_comparison.llama1B}]{
        \includegraphics[width=0.5\linewidth]{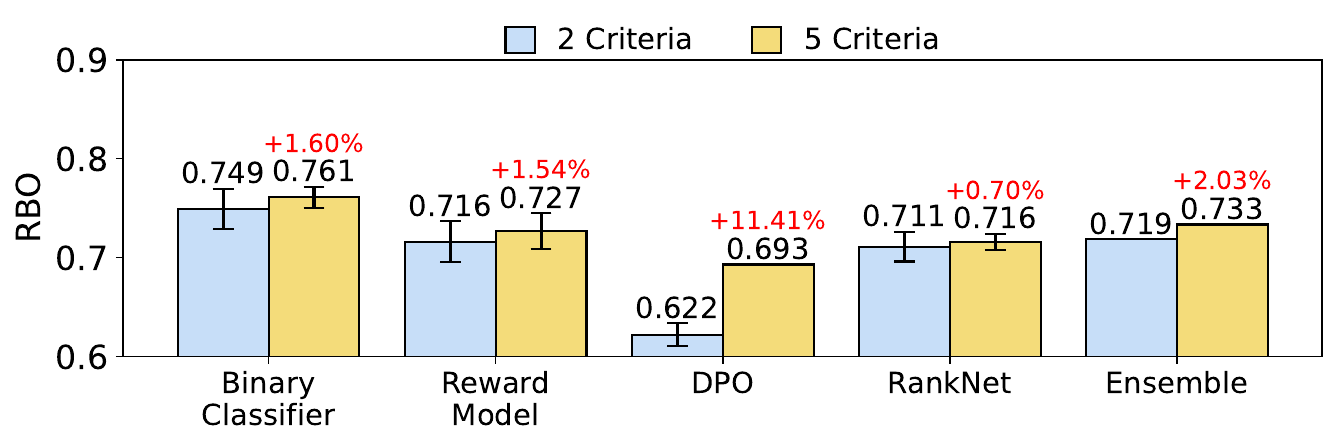}}
    \subfloat[Llama-1B-IT\label{fig:criteria_comparison.llama1B-IT}]{
        \includegraphics[width=0.5\linewidth]{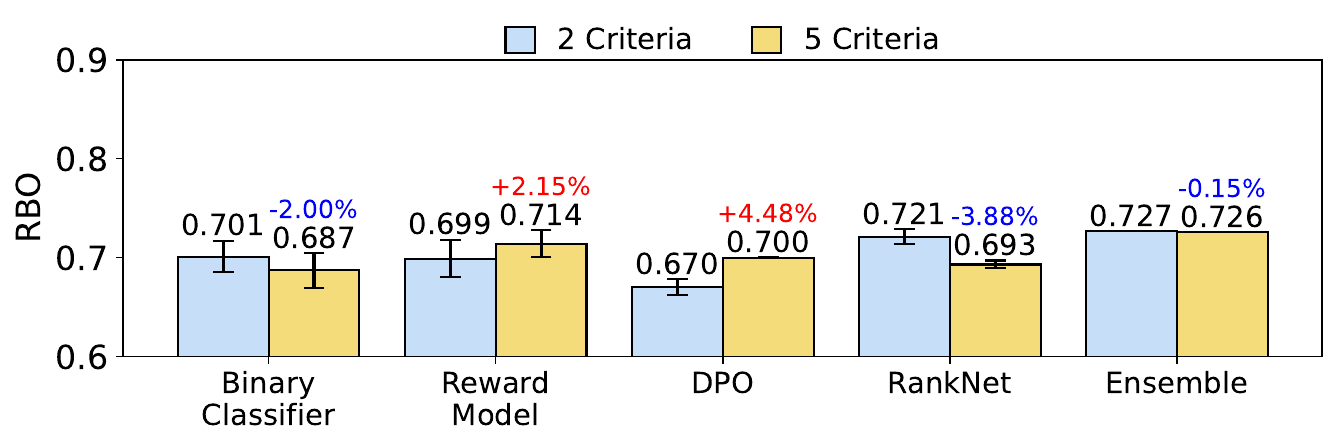}}

    \subfloat[Llama-3B\label{fig:criteria_comparison.llama3B}]{
        \includegraphics[width=0.5\linewidth]{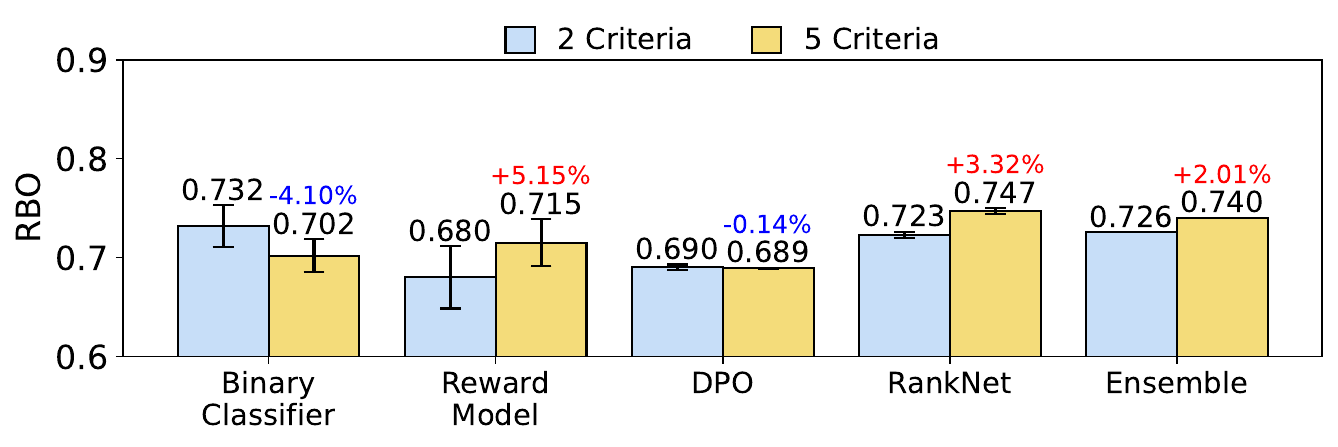}}
    \subfloat[Llama-3B-IT\label{fig:criteria_comparison.llama3B-IT}]{
        \includegraphics[width=0.5\linewidth]{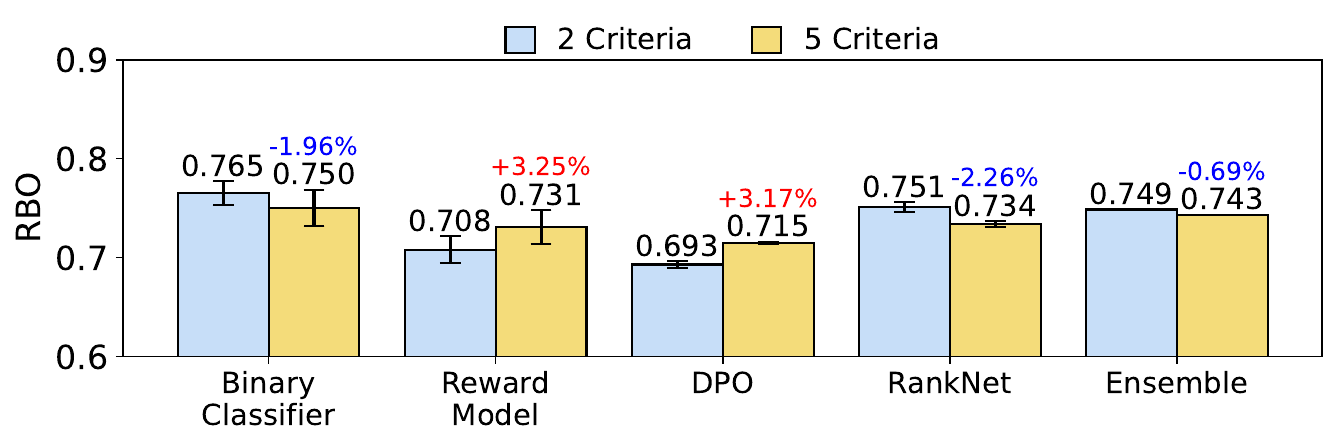}}

    \subfloat[Qwen-3B-IT\label{fig:criteria_comparison.qwen3B-IT}]{
        \includegraphics[width=0.5\linewidth]{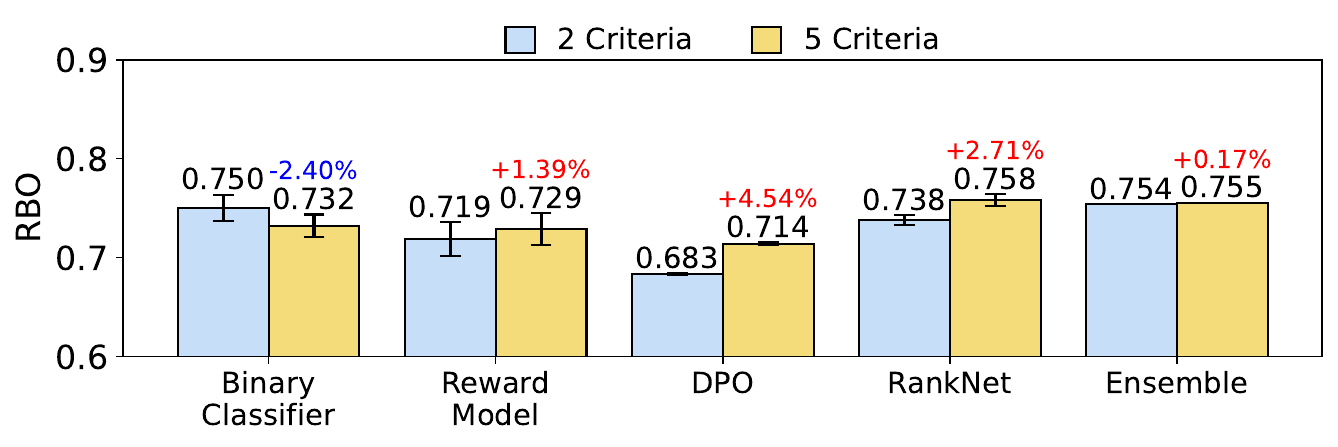}}

    \caption{\label{fig:criteria_comparison} Overall performance across varying numbers of feedback criteria.}
\end{figure*}

\subsection{Performance Analysis by DM Ratio in the DA$\rightarrow$DM Scenario}\label{sec:experimental_results:ratio}

We analyzed how varying the proportion of DM (5-100\%) in the DA$\rightarrow$DM scenario affects model performance. Figure \ref{fig:performance_ratio_filtered} presents the results for Llama-3B-IT and Qwen-3B-IT, the models that achieved the strongest performance under most approaches.

For Llama-3B-IT, the binary classifier, DPO, and Ensemble outperformed the DM$\rightarrow$DM  even with only 5\% of human-annotated DM and DA data. Similarly, Reward Model and RankNet exceeded DM$\rightarrow$DM performance within the 5–10\% annotation range. In contrast, Qwen-3B-IT surpassed DM$\rightarrow$DM primarily within the 50–75\% or 75–100\% annotation ranges. Although Qwen-3B-IT is not as efficient as Llama-3B-IT, the results suggest that high performance can be achieved with minimal human annotation costs. The overall performance of Llama-1B, Llama-1B-IT, and Llama-3B models is illustrated in Figure \ref{fig:performance_ratio} (see Appendix \ref{appendix:additional_experimental-results}).

\subsection{Performance Analysis by Number of Feedback Criteria}\label{sec:experimental_results:compare_feedback_criteria}
To investigate the impact of feedback criteria, we compared two training configurations: using all five criteria versus using only two essential criteria (Correct and Revealing). For the experiments, we generated an additional version of DG that includes only two feedback criteria and trained a ranking model in the DG$\rightarrow$DM scenario.

Figure \ref{fig:criteria_comparison} shows that increasing the number of feedback criteria from two to five consistently improves Llama-1B across all approaches, with DPO exhibiting the largest gain (+11.41 \%). For Qwen-3B-IT, every approach except the binary classifier benefits from the richer feedback, and the remaining models display improvements in selected approaches. These results suggest that incorporating richer feedback information enhances model generalization.

\section{Conclusion}\label{sec:conclusion}
In this study, we proposed the \textbf{F}eedback Dataset Generation Framework for \textbf{E}nglish \textbf{A}I \textbf{T}utoring (FEAT), which utilizes LLMs to generate teacher feedback and build preference datasets for English tutoring. We evaluated ranking models on three datasets—DIRECT-Manual (DM), DIRECT-Generated (DG), and DIRECT-Augmented (DA)—constructed via FEAT.

Results showed that models based on DG performed competitively with DM-based models. Moreover, supplementing with only 5–10\% human-annotated DM led to superior performance than using the full DM dataset. These findings demonstrate that high performance can be achieved with minimal human effort with our FEAT framework.
In future research, we will extend our framework to broader educational scenarios.

\section*{Limitations}
In this study, we explored the feasibility of LLM-based teacher feedback generation and preference dataset construction using the FEAT framework. However, the study has the following limitations:  

First, while we constructed an English tutoring scenario using the MCTest dataset, further research is required to assess the generalizability of the framework across diverse educational datasets. 

Second, we only conducted the ranking model experiments using 1B and 3B LLMs. Future work should explore the applicability of larger LLMs (e.g., 7B, 13B, 70B) to evaluate their impact on ranking performance.

Third, we employed a pairwise approach for ranking model training. We plan to explore preference dataset construction and training strategies applicable to pointwise and listwise ranking approaches in future research.

\section*{Acknowledgement}
This work was supported by Institute of Information \& Communications Technology Planning \& Evaluation (IITP) grant funded by the Korea government (MSIT) (2019-0-00004, Development of semisupervised learning language intelligence technology and Korean tutoring service for foreigners), the National Research Foundation of Korea (NRF) grant funded by the Korea government (MSIT) (No. RS-2025-0055621731482092640101), and research fund of Chungnam National University.

\bibliography{refer}
\bibliographystyle{acl_natbib}

\appendix
\section{Related Works}\label{sec:related_works}
\subsection{Feedback in Education}\label{sec:related_works:feedback_in_education}
In the field of education, teacher feedback plays a crucial role in enhancing students' learning experiences and achievements. In particular, immediate and appropriate feedback positively impacts students' cognitive, emotional, and motivational outcomes \cite{education_feedback}.

Research has been conducted on designing effective feedback strategies. \citet{education_feedback2} proposed seven principles for effective feedback, while \citet{education_feedback3} analyzed the impact of feedback on learning and investigated its key components. \citet{feedback_criteria1, feedback_criteria2} proposed five criteria for evaluating feedback quality, designing them to help students understand clear directions for improvement and maintain motivation. Additionally, research applying feedback criteria has been conducted in fields such as programming education \cite{feedback_evaluation1}.

\subsection{Large Language Models in Education}\label{sec:related_works:llm_in_education}
Advancements in large language models (LLMs) have significantly impacted the field of education \cite{llm_education1,llm_education2,llm_education3,llm_education4,llm_education5}. The integration of LLMs with educational technology has been applied across various domains, including automated short answer grading \cite{automated_short_answer_grading}, automated essay scoring \cite{automated_essay_scoring}, automated distractor generation \cite{automated_distractor_generation1, automated_distractor_generation2}, and automatic question generation \cite{question_generation1,question_generation2,question_generation3,question_generation4,question_generation5}.

Research has also been conducted on automated feedback systems to provide better feedback to students \cite{llm_education5,feedback_generation1}. Additionally, LLM-powered personalized feedback generation has contributed to reducing teachers’ workloads and improving the efficiency of online education \cite{feedback_generation3}. Beyond feedback generation, LLMs have also been utilized for feedback quality assessment. Studies have proposed LLM-based feedback generation and evaluation frameworks in domains such as programming assignments \cite{feedback_evaluation1} and mathematics \cite{feedback_criteria2}, demonstrating the potential of LLMs in educational settings.

\subsection{Learning to Rank Approaches}\label{sec:related_works:learning_to_rank}
Learning to Rank (LTR) is widely used in information retrieval and recommendation systems, aiming to learn the optimal ranking of items for a given query. LTR methodologies are generally categorized into three approaches: Pointwise, Pairwise, and Listwise.

Pointwise approaches, such as MCRank \cite{mcrank} and PRank \cite{prank}, predict a relevance score for each item individually and rank them based on these scores. Pairwise approaches learn the relative preference between two items, with representative algorithms including RankNet \cite{ranknet}, LambdaRank \cite{lambdarank}, RankSVM \cite{ranksvm}, RankBoost \cite{rankboost}, GBRank \cite{gbrank}, and FRank \cite{frank}. Listwise approaches consider the entire item list as a single input and optimize its order holistically, with prominent algorithms such as ListNet \cite{listnet}, ListMLE \cite{listmle}, and SoftRank \cite{softrank}.

Recent LTR research has expanded to leverage LLMs for ranking tasks \cite{ultrafeedback}. \citet{llm_pairwise_ranking, prp-graph} proposed LLM-based pairwise ranking methods, demonstrating the potential of large-scale language models in ranking optimization.

\begin{table}
\centering
\begin{adjustbox}{max width=\linewidth}
\begin{tabular}{cccccc}
\toprule 
 & \textbf{Human} & \textbf{DIRECT} & \textbf{PrepTutor} & \textbf{GPT-3.5} & \textbf{GPT-4}\tabularnewline
\midrule 
Word & 10.02 & 9.27 & 29.37 & 17.84 & 21.01\tabularnewline
Token & 13.98 & 13.25 & 36.60 & 22.61 & 26.58\tabularnewline
\bottomrule
\end{tabular}
\end{adjustbox}
\caption{\label{tab:direct_m_feedback_statistics}Average feedback word length and token length in the DIRECT-Manual dataset.}
\end{table}

\begin{table}
\centering
\begin{adjustbox}{max width=\linewidth}
\begin{tabular}{cccc}
\toprule 
 & \multirow{2}{*}{\textbf{\# Data}} & \multicolumn{2}{c}{\textbf{Average Words Per:}}\tabularnewline
\cmidrule{3-4} 
 &  & \textbf{Story} & \textbf{Question}\tabularnewline
\midrule 
\multicolumn{1}{c}{DIRECT-Manual} & 5,500 & 193.05 & 11.90\tabularnewline
MCTest & 1,480 & 202.71 & 7.79\tabularnewline
\bottomrule
\end{tabular}
\end{adjustbox}
\caption{\label{tab:dataset_statistics}DIRECT-Manual and MCTest dataset statistics.}
\end{table}

    \section{Details of DIRECT-Manual Dataset}\label{appendix:details-of-direct_m}
\subsection{Feedback Generation Process}
The DIRECT-Manual (DM) consists of five feedback candidates, each generated through different methods. The details of these candidates are as follows:

\begin{itemize}
    \setlength{\itemsep}{0pt} 
    \setlength{\parsep}{0pt}  
    \item \textbf{Human}: Feedback written by human annotators.
    \item \textbf{DIRECT}: Feedback generated using GPT-2 trained on the DIRECT dataset.
    \item \textbf{PrepTutor}: Feedback generated using GPT-2 fine-tuned on external domain-specific feedback data.
    \item \textbf{GPT-3.5}: Feedback generated using GPT-3.5-turbo-0613.
    \item \textbf{GPT-4}: Feedback generated using GPT-4-0613.
\end{itemize}

\subsection{Feedback Ranking Process}
The feedback candidates were ranked by human annotators based on two criteria:

\begin{tcolorbox}[colframe=black, colback=gray!10, rounded corners, left=-7pt]
\begin{itemize}
    \setlength{\itemsep}{0pt}
    \item \textbf{Correct}: The feedback provides specific factual information based on the student's response or the given text.
    \item \textbf{Revealing}: The feedback guides the student toward the correct answer without explicitly stating it.
\end{itemize}
\end{tcolorbox}

\begin{figure}
    \centering
    \includegraphics[width=\linewidth]{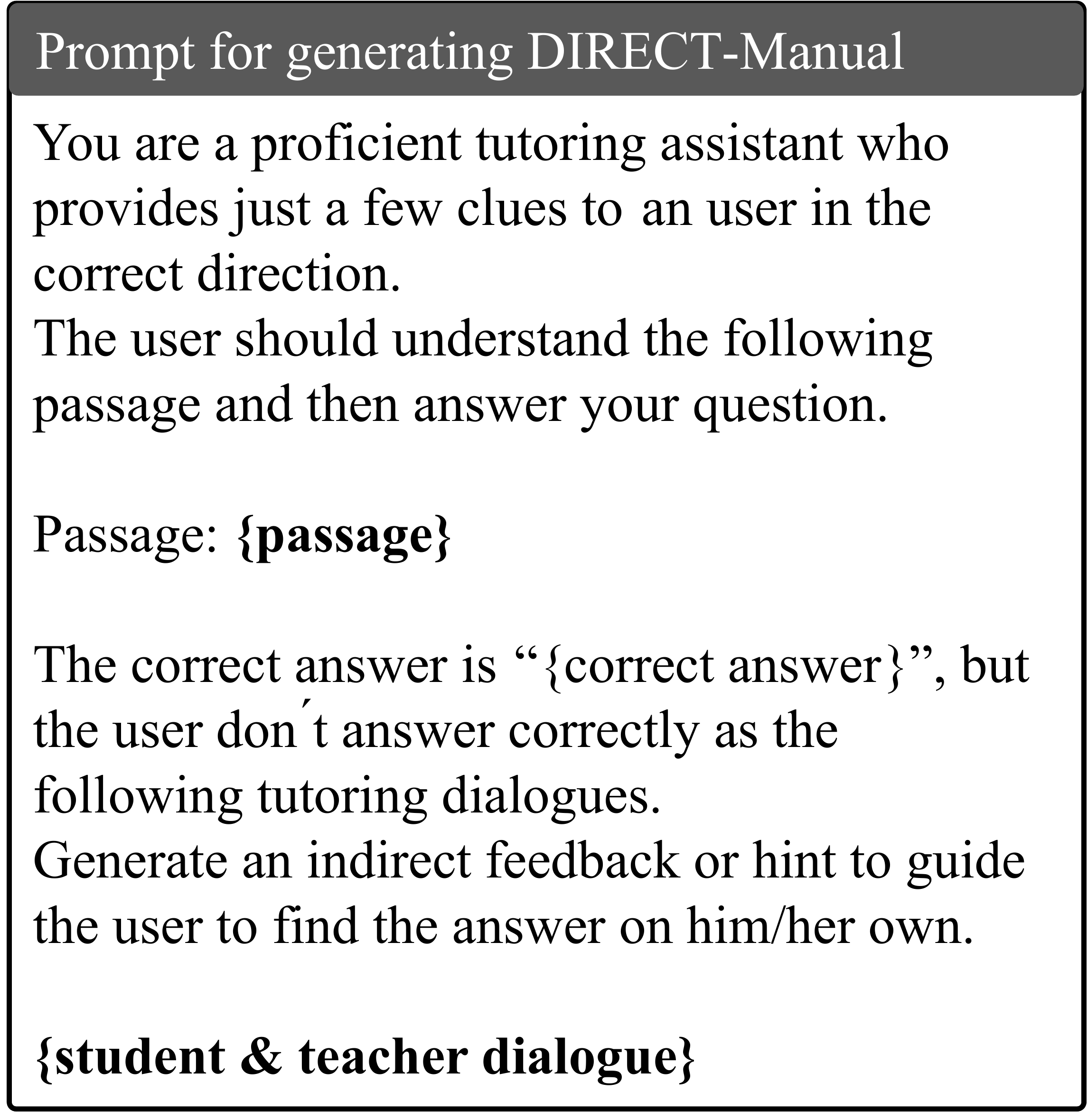}
    \caption{Prompt for generating DIRECT-Manual.}
    \label{fig:prompt_direct_m}
\end{figure}

Table \ref{tab:direct_m_feedback_statistics} presents the average length of feedback candidates, while Table \ref{tab:dataset_statistics} provides overall dataset statistics. Figure \ref{fig:prompt_direct_m} illustrates the prompt used for feedback generation with GPT-3.5 and GPT-4 in the DM.

\section{Details of DIRECT-Generated Dataset}\label{appendix:details-of-direct_a}
\subsection{Dataset Preprocessing Process}
The DIRECT-Generated (DG) dataset was constructed based on MCTest \cite{mctest}, which consists of stories designed for students in grades 1–4, along with corresponding questions and four answer options.

The dataset construction involved the following preprocessing steps:

\begin{enumerate}
    \item The question field from MCTest was used as the teacher's question.
    \item The answer field was used as the student's correct response.
    \item One option from the answer choices was randomly selected as the student's incorrect response.
\end{enumerate}

\subsection{Feedback Generation Process}
We utilized LLMs to automatically generate teacher feedback. The prompt for feedback generation was designed to include the story, question, the student's incorrect response, and the correct response. Additionally, the five feedback criteria defined by \citet{hyein-llm-evaluators} were applied to ensure educationally effective feedback. The characteristics of each criterion are as follows:

\begin{tcolorbox}[colframe=black, colback=gray!10, rounded corners, left=-7pt]
\begin{itemize}
    \setlength{\itemsep}{0pt}
    \item \textbf{Correct}: The feedback should be factually accurate and directly related to the student's response and the question. 
    \item \textbf{Revealing}: The feedback should avoid explicitly providing the correct answer to the student.
    \item \textbf{Guidance}: The feedback should offer direction or hints to help the student progress towards the right answer. 
    \item \textbf{Diagnostic}: The feedback should pinpoint and address any misconceptions or errors made by the student.
    \item \textbf{Encouragement}: The feedback should convey a positive and supportive tone to motivate the student. 
\end{itemize}
\end{tcolorbox}

The following LLMs were used for feedback generation:
\begin{itemize}
    \setlength{\itemsep}{0pt}
    \item GPT-4o \cite{gpt4}
    \item Claude-3\footnote{claude-3-5-sonnet-20240620} \cite{claude}
    \item Llama-3.1-70B-Instruct\footnote{\url{https://huggingface.co/meta-llama/Llama-3.1-70B-Instruct}} \cite{llama3}
\end{itemize}

\begin{figure*}
    \centering
    \includegraphics[width=\linewidth]{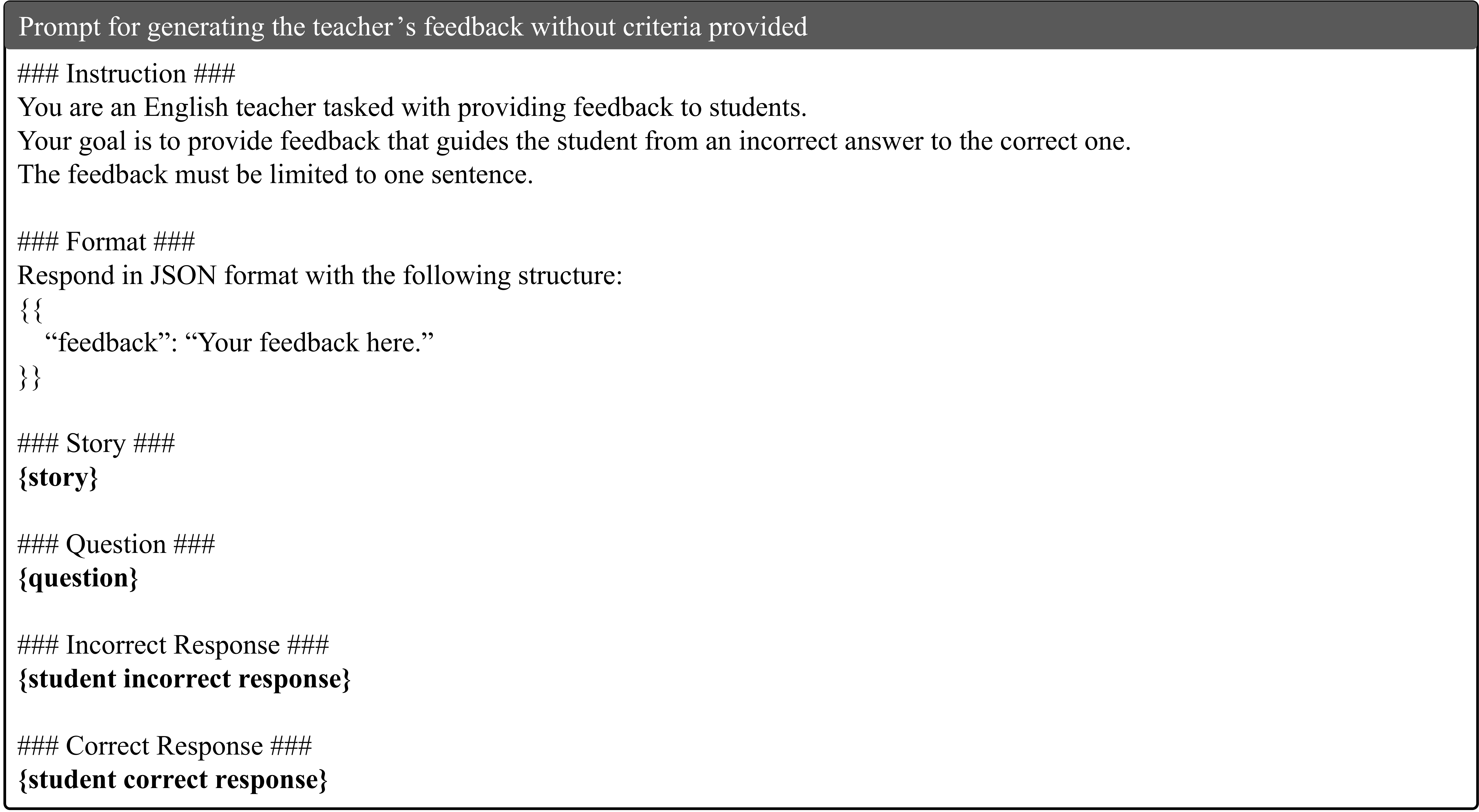}
    \caption{Prompt for generating the teacher’s feedback with criteria provided.}
    \label{fig:prompt_w_criteria}
\end{figure*}

\begin{figure*}
    \centering
    \includegraphics[width=\linewidth]{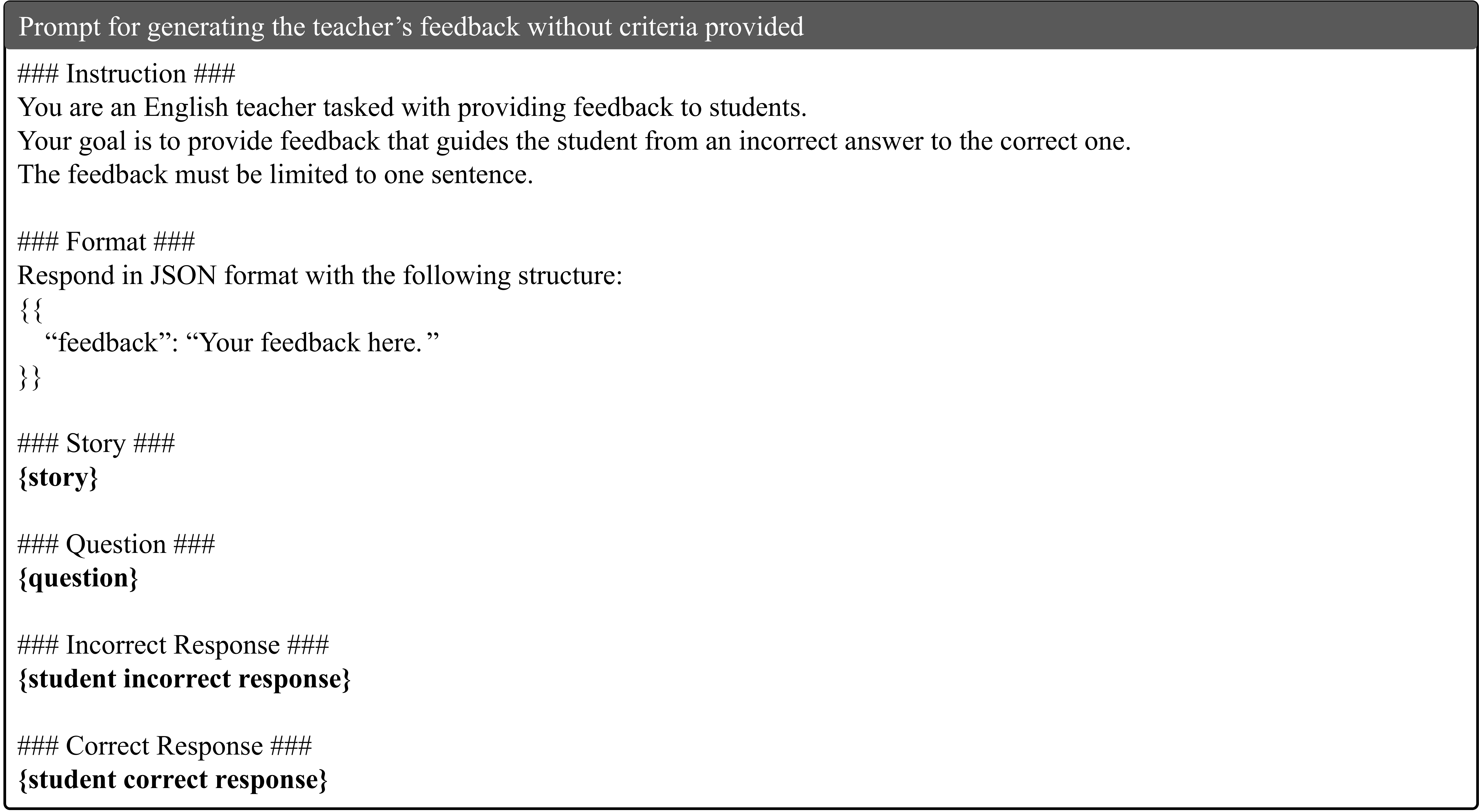}
    \caption{Prompt for generating the teacher’s feedback without criteria provided.}
    \label{fig:prompt_wo_criteria}
\end{figure*}

Figures \ref{fig:prompt_w_criteria} and \ref{fig:prompt_wo_criteria} illustrate example prompts. The prompts were designed to generate teacher feedback that guides students from incorrect to correct responses. The feedback generated by all three LLMs was aggregated and then split into train and test datasets at a 9:1 ratio.

Table \ref{tab:example_feedback_generation} presents examples of teacher feedback generated under different prompt strategies in the DG dataset. Notably, when feedback criteria were not applied (w/o criteria), the generated feedback often explicitly stated the correct answer. In contrast, when feedback criteria were applied (w/ criteria), the generated feedback was more structured and pedagogically aligned.

\section{Implementation Details}\label{appendix:implementation-details}
\begin{figure}
    \centering
    \includegraphics[width=\linewidth]{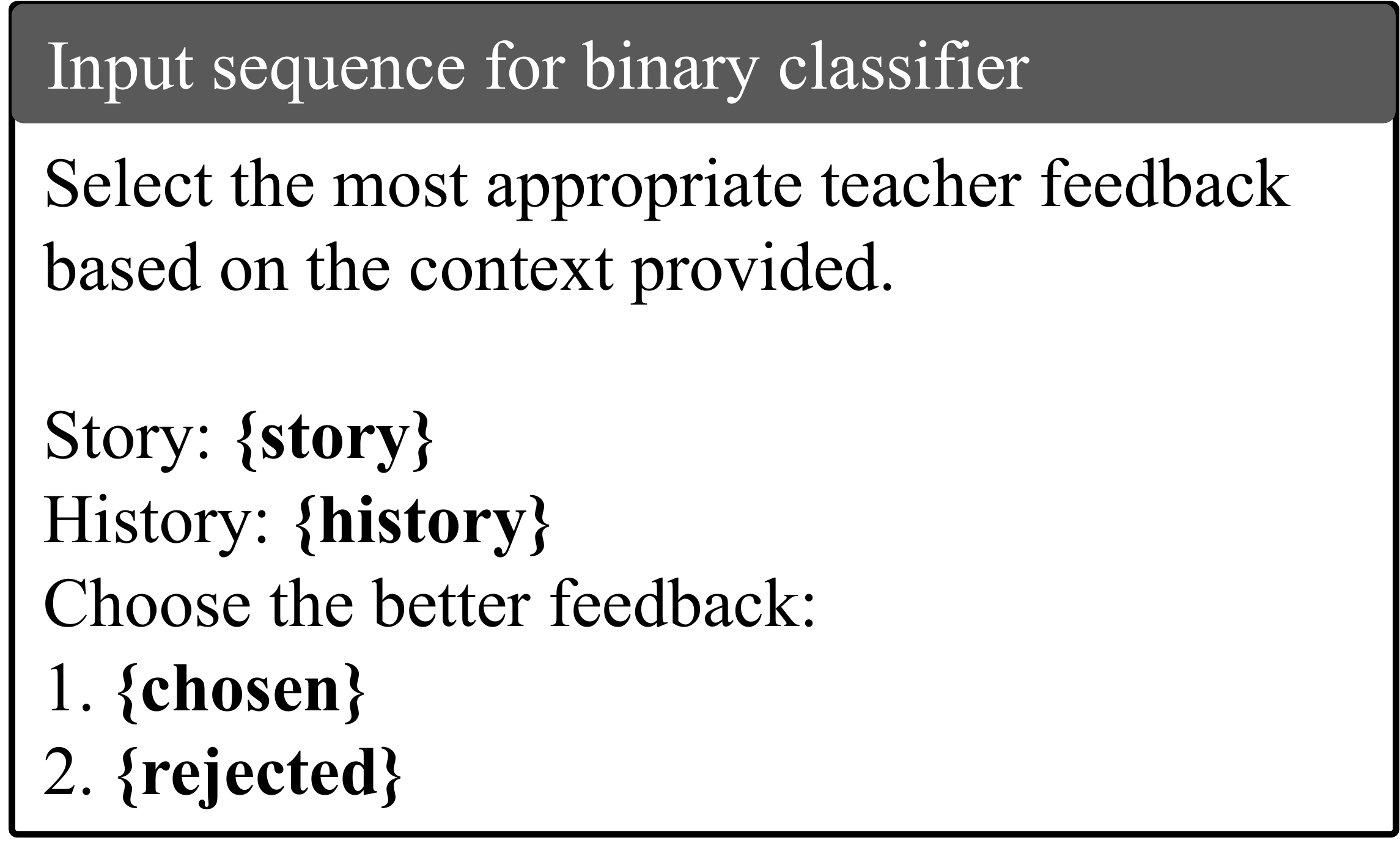}
    \caption{Input data format for binary classifier.}
    \label{fig:input_sequence_binary_classifier}
\end{figure}

\begin{figure}
    \centering
    \includegraphics[width=\linewidth]{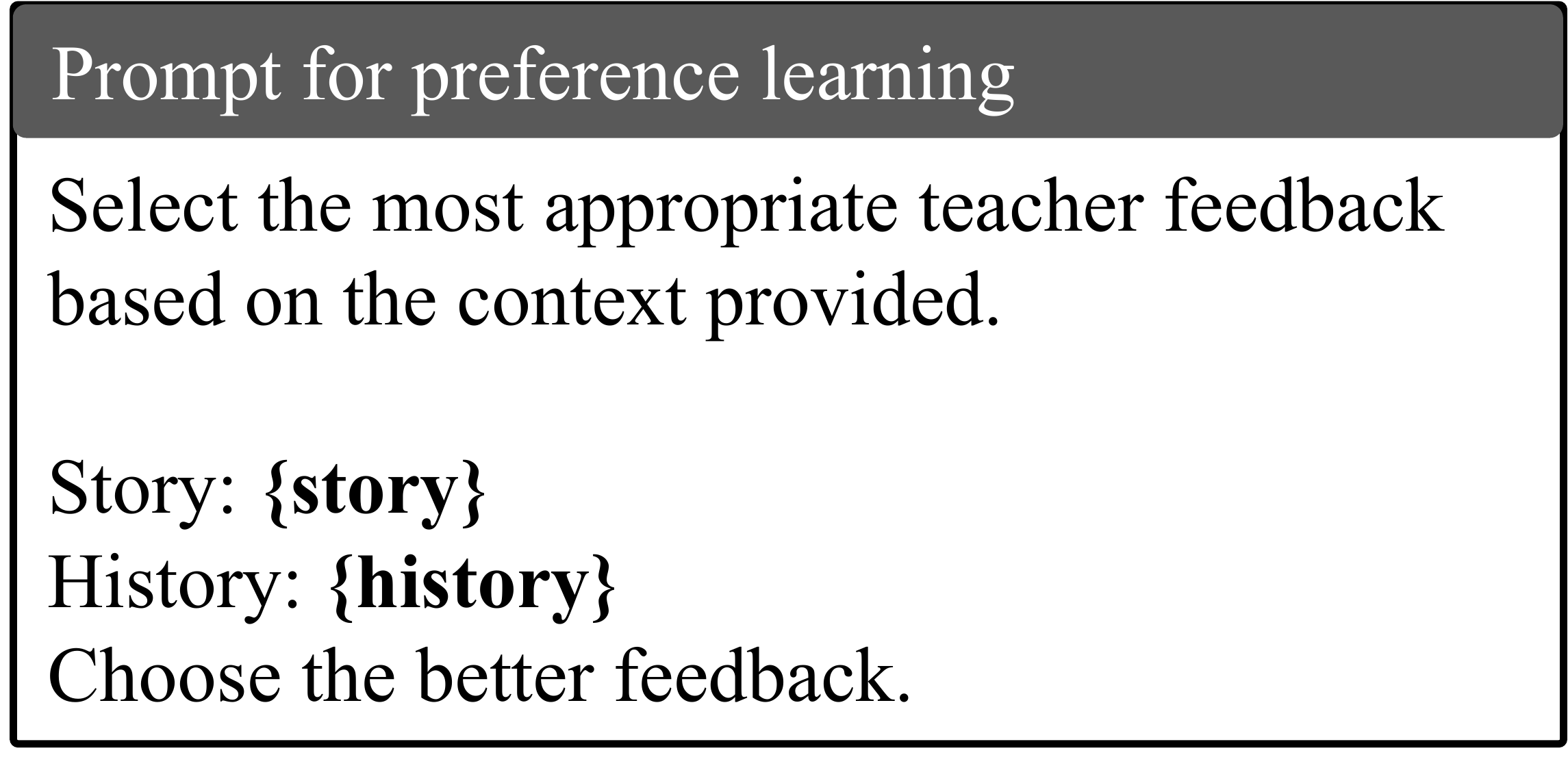}
    \caption{Prompt for reward model, DPO, and RankNet.}
    \label{fig:prompt_preference_learning}
\end{figure}

All experiments were conducted in NVIDIA A100 (40GB VRAM) GPUs and implemented using the PyTorch. The Hugging Face \cite{huggingface} was utilized for model training. All models were fine-tuned using the Low-Rank Adaptation (LoRA) \cite{lora}. The versions of the models used are listed in Table \ref{tab:model_list}, and detailed hyperparameter settings are provided in Table \ref{tab:ranking_model_hyperparameters}. The input format of the ranking model is illustrated in Figure \ref{fig:input_sequence_binary_classifier} and Figure \ref{fig:prompt_preference_learning}.

\section{Additional Experimental Results}\label{appendix:additional_experimental-results}
Table \ref{tab:overall_performance} presents the overall performance across different ranking model approaches. All experiments were conducted five runs with different random seeds.

\begin{figure*}[p]
    \centering
    \subfloat[Llama-1B\label{fig:performance_ratio_llama1B}]{
        \includegraphics[width=\linewidth]{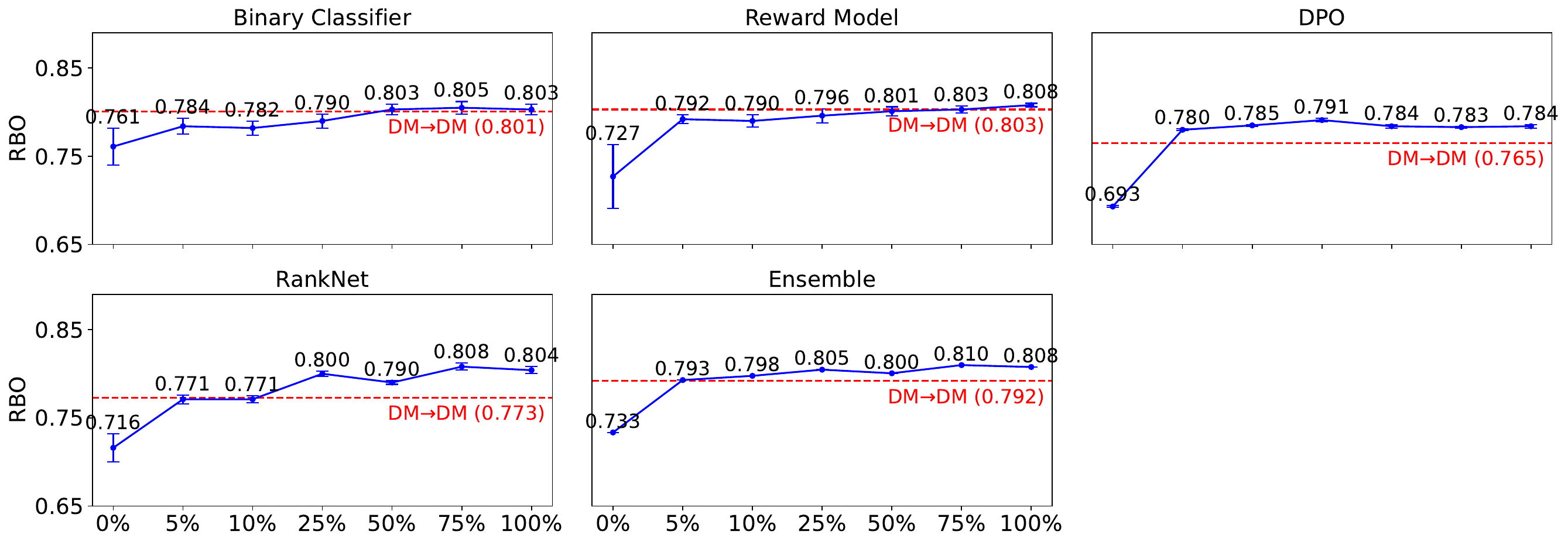}}

    \subfloat[Llama-1B-IT\label{fig:performance_ratio_llama1B-IT}]{
        \includegraphics[width=\linewidth]{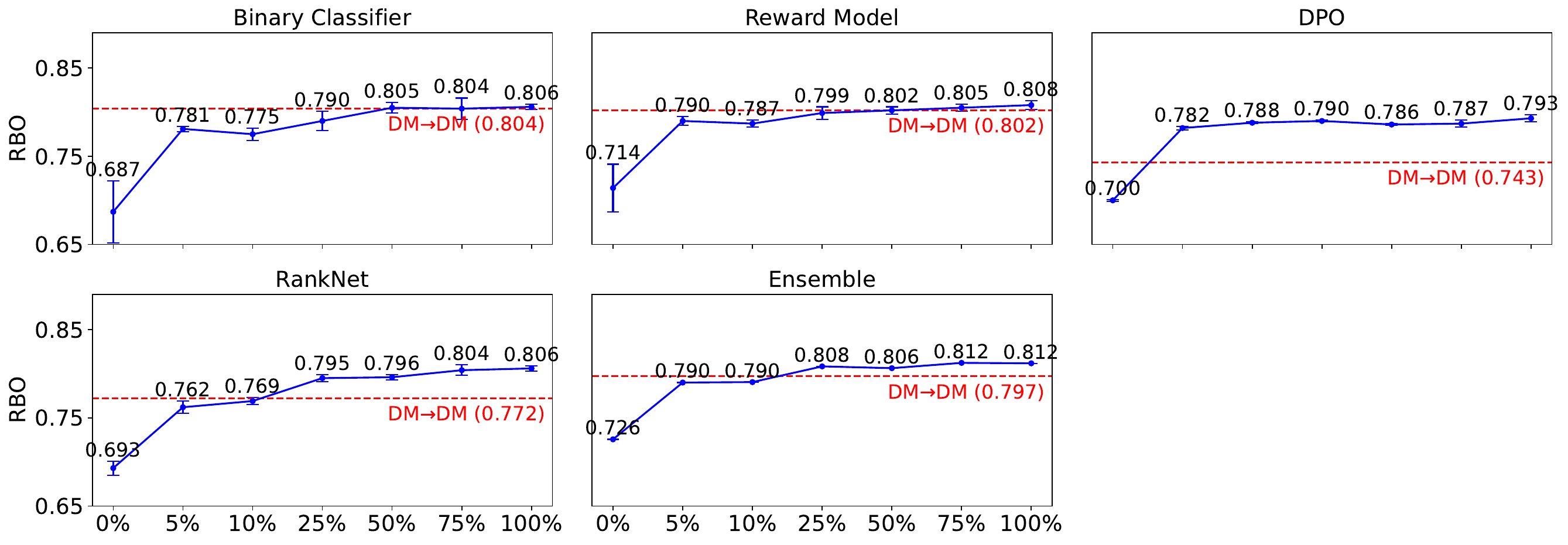}}

    \subfloat[Llama-3B\label{fig:performance_ratio_llama3B}]{
        \includegraphics[width=\linewidth]{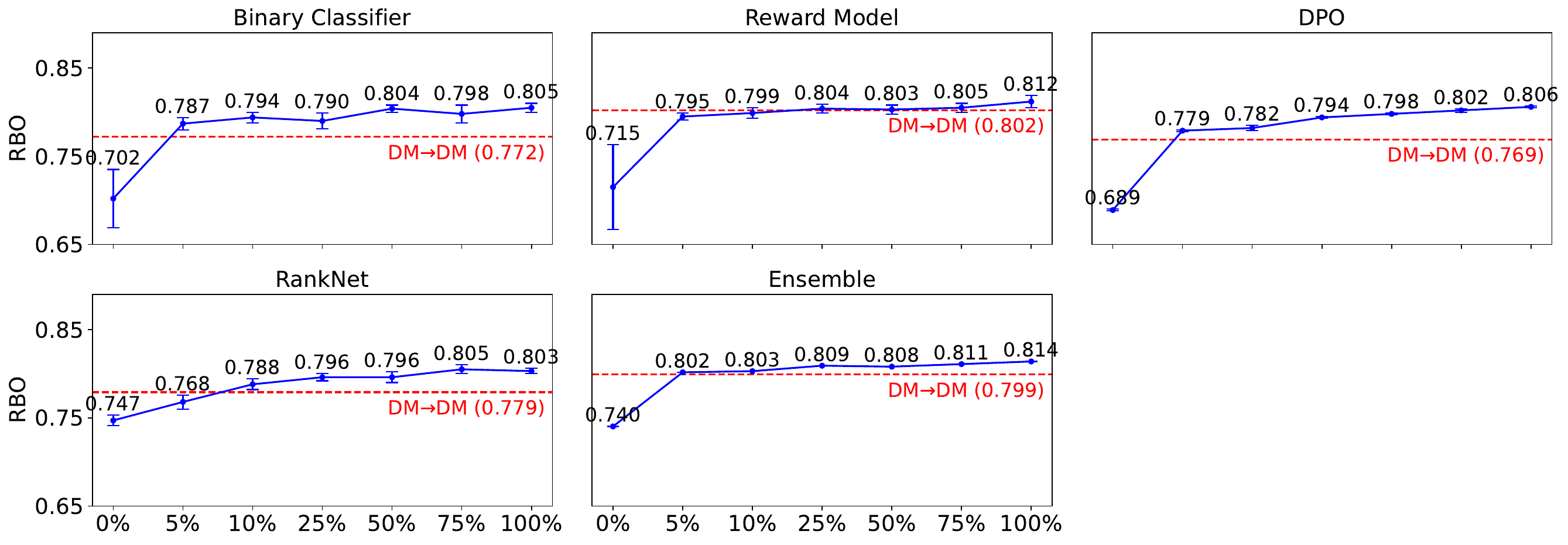}}

    \caption{\label{fig:performance_ratio} Overall performance across varying DM dataset proportions in the DA$\rightarrow$DM scenario. Error bars represent the standard deviation calculated from the results of five different seeds.}
\end{figure*}

Figure \ref{fig:performance_ratio} summarizes the results of the DIRECT-A (DA) $\rightarrow$DIRECT-M (DM) scenario described in Section \ref{sec:experimental_results:ratio} for the Llama-1B, Llama-1B-IT, and Llama-3B. In most ranking model approaches, performance improved as the proportion of DM increased. For every model, DPO exceeds the DM$\rightarrow$DM baseline even when trained with only 0–5\% of the DM data. Notably, Llama-3B surpasses the DM$\rightarrow$DM baseline in all methods with at most 10–25\% of the DM data.

Figures \ref{fig:case_study_best} and \ref{fig:case_study_worst} show examples of high and low RBO scores between ground-truth rankings and predicted rankings, respectively. When the two rankings are nearly identical, the RBO reaches 0.8333; when they diverge markedly, the score drops to 0.4166.

\begin{table*}
\begin{adjustbox}{max width=\linewidth}
\centering
\begin{tabular}{c|c|c}
\toprule 
 & \textbf{Model Name} & \textbf{Generated Teacher Feedback}\tabularnewline
\midrule 
\multirow{7}{*}{w/ criteria} & \multirow{2}{*}{GPT-4o} & Consider the part of the story where Tom and his friends struggled
the most \tabularnewline
 &  & and needed to exert extra effort to complete the task.\tabularnewline
 \cmidrule{2-3}
 & \multirow{2}{*}{Claude-3} & Consider re-reading the part of the story that describes the difficulty
level of fixing each item, \tabularnewline
 &  & paying special attention to which task was described as `very hard'. \tabularnewline
 \cmidrule{2-3}
 & \multirow{3}{*}{Llama-3.1-70B} & Consider re-examining the text for clues about \tabularnewline
 &  & which task Tom and his friends found particularly challenging, \tabularnewline
 &  & as indicated by the use of specific adjectives to describe the difficulty
level of each task. \tabularnewline
\midrule 
\midrule 
\multirow{8}{*}{w/o criteria} & \multirow{2}{*}{GPT-4o} & Remember, the story mentions that fixing the window was very hard, \tabularnewline
 &  & indicating it was the hardest thing for Tom and his friends to fix. \tabularnewline
 \cmidrule{2-3}
 & \multirow{3}{*}{Claude-3} & While the toilet was mentioned first, \tabularnewline
 &  & the story explicitly states that fixing the window was `very hard' \tabularnewline
 &  & and required all of them to push `really hard', indicating it was
the most challenging task.\tabularnewline
\cmidrule{2-3}
 & \multirow{3}{*}{Llama-3.1-70B} & You might want to reconsider your answer, \tabularnewline
 &  & as the passage states that fixing the window was very hard \tabularnewline
 &  & and required a lot of effort from Tom and his friends to open it. \tabularnewline
\bottomrule
\end{tabular}
\end{adjustbox}
\caption{\label{tab:example_feedback_generation}Examples of DIRECT-Generated for each prompt strategy.}
\end{table*}

\begin{table}
\centering
\begin{adjustbox}{max width=\linewidth}
\begin{tabular}{cc}
\toprule 
\textbf{Model Name} & \textbf{Version}\tabularnewline
\midrule 
Llama-1B & Llama-3.2-1B\tabularnewline
Llama-1B-IT & Llama-3.2-1B-Instruct\tabularnewline
Llama-3B & Llama-3.2-3B\tabularnewline
Llama-3B-IT & Llama-3.2-3B-Instruct\tabularnewline
Qwen-3B & Qwen2.5-3B-Instruct\tabularnewline
\bottomrule
\end{tabular}
\end{adjustbox}
\caption{\label{tab:model_list}Model names and versions Used for training the ranking model.}
\end{table}

\begin{table}
\begin{adjustbox}{max width=\linewidth}
\centering
\begin{tabular}{cc}
\toprule 
\textbf{Hyperparameter} & \textbf{Value}\tabularnewline
\midrule 
\multicolumn{2}{c}{\cellcolor{gray!30}\textit{Training Hyperparameters}}\tabularnewline
\midrule 
Learning rate & 5e-05\tabularnewline
Batch size & 8\tabularnewline
Training epochs & 5\tabularnewline
Max sequence length & 1,024\tabularnewline
Random seeds & 0, 42, 500, 1000, 1234\tabularnewline
\midrule 
\multicolumn{2}{c}{\cellcolor{gray!30}\textit{Lora Config}}\tabularnewline
\midrule
Rank & 16\tabularnewline
Alpha & 32\tabularnewline
Dropout & 0.05\tabularnewline
\bottomrule
\end{tabular}
\end{adjustbox}
\caption{\label{tab:ranking_model_hyperparameters}Hyperparameters for training the ranking model.}
\end{table}

\begin{table*}
\centering
\begin{adjustbox}{max width=\linewidth}
\begin{tabular}{ccccccccccccccc}
\toprule 
 & \multicolumn{2}{c}{Classifier} &  & \multicolumn{2}{c}{Reward Model} &  & \multicolumn{2}{c}{DPO} &  & \multicolumn{2}{c}{RankNet} &  & \multicolumn{2}{c}{Ensemble}\tabularnewline
\cmidrule{2-3} \cmidrule{5-6}\cmidrule{8-9} \cmidrule{11-12} \cmidrule{14-15} 
Model Name & DM\textrightarrow DM & DG\textrightarrow DM &  & DM\textrightarrow DM & DG\textrightarrow DM &  & DM\textrightarrow DM & DG\textrightarrow DM &  & DM\textrightarrow DM & DG\textrightarrow DM &  & DM\textrightarrow DM & DG\textrightarrow DM\tabularnewline
\midrule 
Llama-1B & 0.801$\pm$0.006 & \textbf{0.761$\pm$0.021} &  & 0.803$\pm$0.004 & 0.727$\pm$0.036 &  & 0.765$\pm$0.012 & 0.693$\pm$0.001 &  & 0.773$\pm$0.005 & 0.716$\pm$0.016 &  & 0.792 & 0.733\tabularnewline
Llama-1B-IT & 0.804$\pm$0.005 & 0.687$\pm$0.035 &  & 0.802$\pm$0.006 & 0.714$\pm$0.027 &  & 0.743$\pm$0.024 & 0.700$\pm$0.001 &  & 0.772$\pm$0.004 & 0.693$\pm$0.008 &  & 0.797 & 0.726\tabularnewline
Llama-3B & 0.772$\pm$0.056 & 0.702$\pm$0.033 &  & 0.802$\pm$0.005 & 0.715$\pm$0.048 &  & 0.769$\pm$0.024 & 0.689$\pm$0.001 &  & 0.779$\pm$0.004 & 0.747$\pm$0.006 &  & 0.799 & 0.740\tabularnewline
Llama-3B-IT & 0.773$\pm$0.056 & 0.750$\pm$0.037 &  & 0.802$\pm$0.004 & \textbf{0.731$\pm$0.034} &  & 0.770$\pm$0.023 & \textbf{0.715$\pm$0.002} &  & 0.774$\pm$0.007 & 0.734$\pm$0.006 &  & 0.798 & 0.743\tabularnewline
Qwen-3B-IT & 0.806$\pm$0.008 & 0.732$\pm$0.023 &  & 0.802$\pm$0.007 & 0.729$\pm$0.032 &  & 0.779$\pm$0.002 & 0.714$\pm$0.003 &  & 0.791$\pm$0.012 & \textbf{0.758$\pm$0.012} &  & 0.811 & \textbf{0.755}\tabularnewline
\bottomrule
\end{tabular}
\end{adjustbox}
\caption{\label{tab:overall_performance}Performance by ranking model approaches. Best results are highlighted in \textbf{bold}. The $\pm$ represents standard deviation from five results of five different seeds. IT refers to the Instruct model.}
\end{table*}

\begin{figure*}
\begin{adjustbox}{max width=\linewidth}
\begin{tcolorbox}[colframe=black, colback=white, rounded corners]
\textbf{Feedback Candidates}
\begin{itemize}
    \item \textbf{Human}: It costs 20,000 to 30,000 or more dollars if you add all the expenses of a year.
    \item \textbf{DIRECT}: no, the cost of tuition is over us\$10, 000 per year.
    \item \textbf{PrepTutor}: good effort! frankly, a college education is not cheap. but, students who attend college earn many years of free education. how much will a college student spend on his or her education?
    \item \textbf{GPT-3.5}:That's close, but it's actually higher than that. Can you think of any other expenses that students have to pay for?
    \item \textbf{GPT-4}: That's a good start, but remember, the \$10,000 is just the average cost of tuition. There are other expenses a student needs to consider like books, housing, and personal expenses. Try to add all these up for a full picture of the annual cost.
\end{itemize}
\hrule
\vspace{0.5em}
\textbf{Ground-truth Ranking}: GPT-4 > GPT-3.5 > DIRECT > Human > PrepTutor\\
\textbf{Predicted Ranking}: GPT-4 > GPT-3.5 > PrepTutor > Human > DIRECT\\
\textbf{RBO}: 0.8833
\end{tcolorbox}
\end{adjustbox}
\caption{An example with a high RBO score between the ground-truth ranking and the predicted ranking.}
\label{fig:case_study_best}
\end{figure*}

\begin{figure*}
\begin{adjustbox}{max width=\linewidth}
\begin{tcolorbox}[colframe=black, colback=white, rounded corners]
\textbf{Feedback Candidates}
\begin{itemize}
    \item \textbf{Human}: You got it half right. There is one more answer.
    \item \textbf{DIRECT}: no, life today has brought new problems.
    \item \textbf{PrepTutor}: good job! actually, life on the planet was not that easy. people had few tools for many years. then came up with idea to create better tools. what is it?
    \item \textbf{GPT-3.5}: That's correct! But there was another reason mentioned in the passage. Can you think of what it might be?
    \item \textbf{GPT-4}: That's correct! Now, can you think about the new problems that life today has brought? Specifically, consider the various types of pollution mentioned in the passage.
\end{itemize}
\hrule
\vspace{0.5em}
\textbf{Ground-truth Ranking}: GPT-3.5 > PrepTutor > Human > GPT-4 > DIRECT\\
\textbf{Predicted Ranking}: GPT-4 > GPT-3.5 > DIRECT > PrepTutor > Human \\
\textbf{RBO}: 0.4166
\end{tcolorbox}
\end{adjustbox}
\caption{An example with a low RBO score between the ground-truth ranking and the predicted ranking.}
\label{fig:case_study_worst}
\end{figure*}

\end{document}